\documentclass[10pt,twocolumn,letterpaper]{article}

\usepackage{cvpr}
\usepackage{xcolor}
\usepackage{enumitem}
\usepackage{multirow}
\usepackage{array}

\setlist{nosep,leftmargin=*}

\definecolor{cvprblue}{rgb}{0.21,0.49,0.74}
\usepackage[pagebackref,breaklinks,colorlinks,allcolors=cvprblue]{hyperref}

\usepackage{booktabs}
\usepackage{url}
\usepackage{tcolorbox}
\usepackage{listings}
\usepackage{subcaption}
\usepackage{xcolor}
\usepackage{algorithm}
\usepackage{algorithmic}
\usepackage{times}
\usepackage{epsfig}
\usepackage{amsmath}
\usepackage{amssymb}
\usepackage{caption}
\usepackage{multirow}
\usepackage[flushleft]{threeparttable}
\usepackage{pifont}
\usepackage{array}
\usepackage{makecell}

\newlength\savewidth
\usepackage[utf8]{inputenc}
\usepackage[T1]{fontenc}
\usepackage{amsfonts}
\usepackage{nicefrac}
\usepackage{microtype}

\usepackage{tabularray}
\usepackage{colortbl}
\usepackage{arydshln}

\definecolor{mygray}{gray}{.9}
\definecolor{mygreen}{rgb}{0, 0.6, 0}
\definecolor{myteal}{rgb}{0.16, 0.47, 0.56}
\definecolor{mypink}{rgb}{0.81, 0.25, 0.44}

\usepackage{soul}
\definecolor{highlightgray}{gray}{0.85}
\sethlcolor{highlightgray}

\def\confName{CVPR}
\def\confYear{2026}

\title{
VISTA: Video Interaction Spatio-Temporal Analysis Benchmark
}

\author{
Alejandro Aparcedo$^{1}$ \quad
Akash Kumar$^{1}$ \quad
Aaryan Garg$^{2}$ \quad
Dalton Pham$^{3}$ \\
Wen-Kai Chen$^{1}$ \quad
Anirudh Bharadwaj$^{1}$ \quad
Aman Chadha$^{4}$\thanks{Work done outside role at Amazon.} \quad
Yogesh Rawat$^{1}$ \\
\\
$^{1}$University of Central Florida \quad
$^{2}$BITS Pilani \\
$^{3}$Ho Chi Minh City University of Science \quad
$^{4}$Amazon GenAI \\
Project Page: \url{https://aaparcedo.github.io/VISTA/}
}

\begin{document}
\maketitle
\begin{abstract}
    Existing benchmarks for Vision-Language Models (VLMs) primarily evaluate spatio-temporal understanding on simple single-action videos, closed attribute sets and
    restricted entity types, failing to capture the freeform, multi-action
    interactions between diverse entities which characterize real-world video
    understanding. Furthermore, the lack of a systematic framework for analyzing
    model failures across complementary spatio-temporal axes hinders
    comprehensive evaluation. To address these gaps, we introduce \textbf{VISTA},
    a \underline{\textbf{V}}ideo \underline{\textbf{I}}nteraction \underline{\textbf{S}}patio-\underline{\textbf{T}}emporal
    \underline{\textbf{A}}nalysis benchmark designed for open-set, multi-entity and
    multi-action spatio-temporal understanding in VLMs. VISTA decomposes videos
    into interpretable entities, their associated actions, and relational
    dynamics, enabling multi-axis diagnostics and unified assessment of relational,
    spatial, and temporal understanding. Our benchmark integrates multiple
    datasets into a single interaction-aware taxonomy and comprises
    \textasciitilde12K curated video-query pairs spanning diverse scenes and
    complexities. We systematically evaluate 11 state-of-the-art VLMs on VISTA,
    and break down aggregate performance across our taxonomy to reveal shortcomings and
    pronounced spatio-temporal biases obscured by traditional metrics. By providing detailed, taxonomy-driven diagnostics on a challenging dataset, VISTA offers a nuanced framework to guide advances in model design, pretraining strategies, and evaluation protocols. Overall, VISTA is the first large-scale, interaction-aware diagnostic benchmark for spatio-temporal understanding in VLMs.
\end{abstract}
    
\section{Introduction}
\label{sec:intro}

Real-world video understanding requires reasoning about complex interactions among entities over time. From pedestrian–vehicle dynamics in autonomous driving to human–human and human–object interactions in surveillance. To achieve this, intelligent visual systems must determine \textit{which} entities interact, \textit{how} they interact, and \textit{where} and \textit{when} these interactions occur. This capability, broadly referred to as spatio-temporal understanding \cite{madan2024foundation,schiappa2023self}, extends beyond traditional object detection and motion analysis, requiring the joint modeling of spatial structure, temporal evolution, and inter-entity relationships.

\begin{figure}[t!]
    \centering
    \includegraphics[width=\columnwidth]{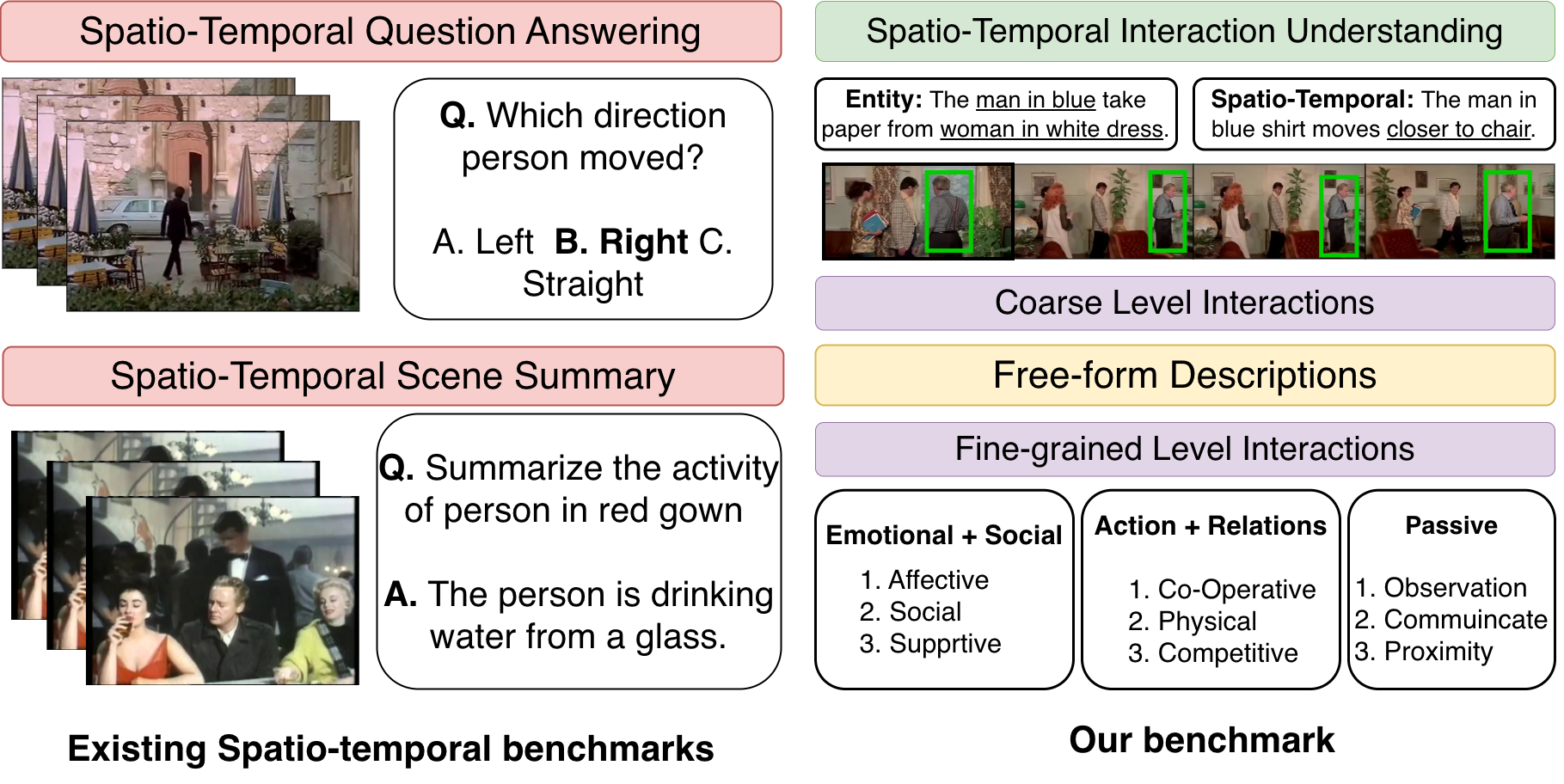}
    \caption{\textbf{VISTA vs. Existing Spatio-Temporal Benchmarks.} Existing benchmarks
    focus on coarse, single-step spatio-temporal understanding without localization. VISTA utilizes grounded evaluation and enables detailed analysis of multi-entity, multi-action dynamics through coarse-to-fine categorization.}
    \label{fig:teaser}
\end{figure}

Vision-Language Models (VLMs)~\cite{uninext-lvlm, liu2024gdino, bai2023qwenvl, chen2023minigptv2, maaz2024videochatgpt, lin2024videollava, ahmad2025t2l} have significantly advanced spatio-temporal understanding by scaling architectures capable of jointly modeling visual and linguistic information. Early evaluation of these models relied on high-level VQA-style benchmarks, which were instrumental in measuring general capabilities. However, subsequent analyses~\cite{feng2025vbenchcomp, goyal2017VQAv2, schiappa2022robustness, grover2024navigating, bai2024hallucination} indicate that performance on such benchmarks can be confounded by linguistic priors limiting their ability to faithfully assess visual understanding. In response, the community has shifted toward grounded benchmarks that validate visual understanding through localization. Early efforts centered on tasks such as object tracking and action recognition, while more recent works~\cite{zhou2025vlm4d, hannan2025svagbench, ahmad2025videomolmo, xu2025mcbench, yuan2025videorefer} have introduced increasingly complex tasks involving multi-entity tracking and 4D reasoning reflecting a growing emphasis on capturing the relational dynamics underlying real-world spatio-temporal understanding. Despite this rapid progress, key limitations remain: existing benchmarks largely reduce performance to aggregate metrics, providing little insight into \textit{where} and \textit{why} models fail. Moreover, as model families expand, the lack of a structured evaluation framework renders consistent, fine-grained cross-model analysis increasingly intractable.

To address these limitations, we introduce \textbf{interaction} as a unifying lens for structured evaluation. Through a systematic dataset aggregation and annotation pipeline, VISTA transforms video-query pairs into a coarse-to-fine interaction-centric representation, factorized into involved entities, spatio-temporal type, and fine-grained interaction type. An overview of the differences between previous work and ours is presented in \autoref{fig:teaser}. Our interaction-centric framework enables three diagnostic capabilities: (a) exposing hidden failure modes, as interaction-level evaluation surfaces systematic limitations masked by aggregate metrics; (b) characterizing generalization patterns, revealing how model behavior stratifies across interaction types, entity configurations, and query formulations; and (c) uncovering directional biases and tendencies, identifying consistent spatial, temporal, and semantic preferences embedded in modern VLMs.

In summary, VISTA provides the first large-scale, interaction-focused diagnostic benchmark for spatio-temporal understanding in VLMs. Our contributions are threefold:
\begin{enumerate} 
\item \textbf{Interaction-centric diagnostic framework:} We introduce a unified coarse-to-fine evaluation taxonomy that decomposes spatio-temporal grounding into interpretable interaction types enabling principled diagnostics across \textasciitilde12K video-query pairs and 11 diverse models.
\item \textbf{Systematic cross-model analysis:} By aggregating and reorganizing multiple datasets under a common interaction-aware structure, we reveal consistent stratification patterns across model families, exposing how architecture, pretraining breadth, and instruction tuning shape understanding. 
\item \textbf{Bias and failure-mode characterization:} Our analysis uncovers prominent failure modes - same-entity disambiguation, linguistic template preferences, and semantic-intent inflation - offering the first interaction-grounded view of systematic reasoning failures in modern VLMs. 
\end{enumerate}

\section{Related Work}

\noindent \textbf{VLM Benchmarks:} The rapid progress of VLMs has been paralleled by increasingly sophisticated benchmarks designed to probe spatio-temporal understanding \cite{schiappa2023self,madan2024foundation}. From early datasets that evaluate general visual understanding \cite{modi2023occlusions,grover2023revealing,chen2015cococaptions, goyal2017VQAv2, plummer2015flickr30k, zellers2019VCR}, to fine-grained spatio-temporal localization tasks \cite{singh2024semi,rana2025omvid,rawat2025active,kumar2025stable,modi2022video,xie2023DOD, zareian2021OVD, anne2017DiDeMo, gao2017tall}. Unlike general video understanding benchmarks \cite{azad2026streamready,li2024mvbench, zhao2025mmvu, fu2025videomme, mangalam2023egoschema} that assess abstract comprehension, spatio-temporal benchmarks emphasize grounded reasoning. Within the segmentation community, 
MOSE~\cite{MOSE} introduced crowded, heavily occluded scenes where targets frequently disappear and reappear, revealing that state-of-the-art VOS methods are brittle under such conditions. Its successor MOSEv2~\cite{MOSEv2} extends this further with adverse weather, low-light environments, camouflaged objects, and non-physical targets. On the language-guided side, MeViSv2~\cite{MeViSv2} shifts the focus from static-attribute referring expressions to \emph{motion}-based descriptions that require genuine temporal reasoning across frames, supporting multi-target and no-target expressions. In the detection-style grounding setting, Spatio-Temporal Video Grounding (STVG) \cite{vidstg, hcstvg} requires joint localization of entities across space and time from freeform relational queries, with recent efforts \cite{zhou2025vlm4d, hannan2025svagbench, ahmad2025videomolmo, xu2025mcbench, yuan2025videorefer} further broadening this toward 4D reasoning, multi-object grounding, and grounded captioning. Yet across both settings, benchmarks largely reduce performance to aggregate metrics, providing little insight into where and why models fail. While prior diagnostic efforts \cite{lin2024stalign, feng2025vbenchcomp} shed light on performance across coarse-grained spatial and temporal categories, they neglect the intricate interaction semantics that critically influence spatio-temporal behavior. VISTA complements segmentation benchmarks by adopting STVG as its diagnostic probe, enabling structured evaluation of \emph{how} and \emph{why} models fail across diverse interaction types, entity configurations, and query formulations---dimensions that mask-based benchmarks do not directly expose. \\
\noindent \textbf{Spatio-temporal understanding in VLMs: } Early spatio-temporal understanding modeled space and time independently under closed-set conditions - spatial models~\cite{faster_rcnn, carion2020detr} handled object detection within fixed categories while temporal models~\cite{simonyan2014twostream, carreira2017activity} targeted action recognition under constrained label settings. Subsequent work advanced vision-language alignment across both spatial and temporal dimensions - through OVD~\cite{zareian2021OVD, minderer2022OVD} and REC~\cite{Chen_2018REC}, culminating in strong detectors such as GLIP~\cite{li2022GLIP} and Grounding-DINO~\cite{liu2024gdino}, while parallel progress in Moment Localization~\cite{anne2017DiDeMo, gao2017tall} enabled language-guided temporal understanding \cite{ahmad2025t2l}. The integration of LLMs into VLMs~\cite{li2023blip2, liu2023llava, bai2023qwenvl} further strengthened multimodal grounding, and video-centric extensions~\cite{azad2026streamready, azad2025hierarq, zhang2023videollama, lin2024videollava} introduced joint spatial-temporal understanding. Despite evaluating increasingly complex spatio-temporal tasks, existing benchmarks reduce performance of VLMs to aggregate metrics, leaving the failure modes, biases, and systematic tendencies of modern VLMs largely undiagnosed.

\section{The VISTA Benchmark}
\label{sec:overview}

\begin{table*}[t]
	\centering
	\renewcommand{\arraystretch}{1.06}
	\scalebox{0.9}{
		\begin{tabular}{l cc ccc cc cccccccc}
				\rowcolor{mygray} 
				\specialrule{1.5pt}{0pt}{0pt}
 	 \cellcolor{mygray} Approach &
 	 \multicolumn{2}{c}{\cellcolor{mygray} Encoders} &
 	 \multicolumn{3}{c}{\cellcolor{mygray} VISTA} & 
 	 \multicolumn{2}{c}{\cellcolor{mygray} Spatio-Temp. } &
 	 \multicolumn{8}{c}{\cellcolor{mygray} Entity } \\ 
 	 	 \rowcolor{mygray} &  Image & Text  & R & F & R\&F & S & T & AA & AO & HA & HH & HO & HS & NI & OO \\ 
 	 \midrule
 	 \multicolumn{16}{c}{\textit{Foundation Model w/o LLMs}} \\ 
 	 \midrule 
 	 GDINO \cite{Liu2023GroundingDM} & Swin-T & BERT & 37.79 & 32.34 & 34.64 & 35.0 & 30.8 & 12.6 & 38.8 & 52.1 & 29.9 & 37.0 & 39.4 & 41.3 & 38.3 \\ 
 	 \midrule
 	 \multicolumn{16}{c}{\textit{Generalist MLLMs}} \\ 
 	 \midrule 
 	 Intern-VL 2.5 \cite{internvl} & InternViT-300M & InternLM2-7B$^{\dagger}$ & 51.11 & 48.65 & 49.73 & 46.3 & \underline{48.0} & 37.9 & 49.8 & 52.2 & \underline{49.2} & 49.5 & 50.9 & 48.2 & 47.7 \\ 
 	 Mini-GPT-v2 \cite{chen2023minigptv2} & EVA-CLIP ViT-G/14 & LLaMA2-7B$^{\dagger}$ & 46.62 & 45.13 & 45.78 & 43.1 & 44.3 & 33.6 & 47.4 & 48.5 & 46.0 & 46.1 & 46.4 & 44.3 & 43.1 \\ 
 	 Sphinx-v2 \cite{sphinx-mllm} & CLIP ViT-L/14 & LLaMA2-7B & 47.79 & 44.28 & 45.82 & 42.6 & 45.0 & 30.4 & 46.9 & 51.0 & 47.0 & 46.8 & 48.1 & 46.0 & 42.5 \\ 
 	 Qwen-VL-Chat \cite{qwenvl-mllm} & ViT-bigG & Qwen-7B & 45.56 & 45.43 & 45.49 & 45.7 & 45.3 & 33.7 & 54.8 & 65.2 & 48.2 & 47.9 & \underline{58.8} & 42.2 & 31.7 \\ 
     Qwen3-VL \cite{qwen3vl} & SigLIP-2 & Qwen3-8B & \textbf{62.85} & \textbf{64.41} & \textbf{63.96} & \textbf{64.8} & \textbf{64.3} & \textbf{59.5} & \textbf{63.2} & \textbf{74.5} & \textbf{66.2} & \textbf{64.7} & \textbf{75.7} & \textbf{60.6} & \textbf{59.1} \\
     MimoVL \cite{coreteam2025mimovltechnicalreport} & Qwen2.5-ViT & MiMo-7B-Base & 43.34 & 42.13 & 44.54 & 36.9 & 43.5 & 40.4 & 38.3 & 38.3 & 45.1 & 36.1 & 46.0 & 43.0 & 27.0 \\
 	 \midrule
 	 \multicolumn{16}{c}{\textit{Specialist MLLMs}} \\ 
 	 \midrule 
 	 Shikra \cite{shikra} & CLIP-ViT-L/14 & Vicuna-1/7B & 30.91 & 31.44 & 31.21 & 29.9 & 32.4 & 20.0 & 28.9 & 36.0 & 34.0 & 35.3 & 38.8 & 31.4 & 24.6 \\ 
 	 Ferret-v1 \cite{ferret} & CLIP-ViT-L/14 & Vicuna-1.3/7B & 17.74 & 22.71 & 20.53 & 20.9 & 23.8 & 14.9 & 23.7 & 23.3 & 26.4 & 24.5 & 33.6 & 19.7 & 13.4 \\ 
 	 CogVLM$^{\ddagger}$ \cite{cogvlm} & EVA2-CLIP-E & Vicuna-1.5/7B & \underline{60.56} & \underline{50.13} & \underline{54.70} & \underline{57.5} & 45.7 & \underline{48.1} & \underline{60.3} & \underline{70.2} & 44.7 & \underline{54.0} & 46.8 & \underline{50.7} & \underline{54.0} \\ 
 	 LLAVA-G \cite{llavag-mllm} & CLIP-ViT-L/14 & Vicuna-1.3/7B & 22.51 & 30.47 & 27.11 & 28.1 & 31.9 & 10.7 & 37.1 & 56.1 & 28.4 & 37.0 & 38.9 & 36.0 & 28.4 \\ 
 	  	 \specialrule{1.5pt}{0pt}{0pt}
		\end{tabular}}
 	 \caption{Main results on VISTA. Referral and freeform query performance is denoted with R and F, respectively. $^{\dagger}$ and $^{\ddagger}$ denotes chat and grounding versions, respectively. \textbf{Bold} and \underline{underline} indicate the best and second best results, respectively.}
	\label{tab:zs_ref_vs_free_det}
\end{table*}

\noindent \textbf{Problem Formulation:}
In VISTA, the input comprises a trimmed video 
$V = (v_{1}, v_{2}, \ldots, v_{T})$ with $T$ frames and a descriptive query caption $Q$ that specifies the primary
subject and activity within the video. The objective is to accurately localize the mentioned subject ($A_R$) in all $T$ frames, thereby forming a spatio-temporal tubelet denoted as $A_R = \{ a_r \}_{t_1}^{t_T}$, where $a_r$ represents the bounding-box for the subject in the $r$-th frame.

\subsection{VISTA Taxonomy}
\label{sec:taxonomy}

\begin{figure}[t!]
    \centering
    \begin{subfigure}[b]{0.48\columnwidth}
        \centering
        \includegraphics[width=\linewidth]
        {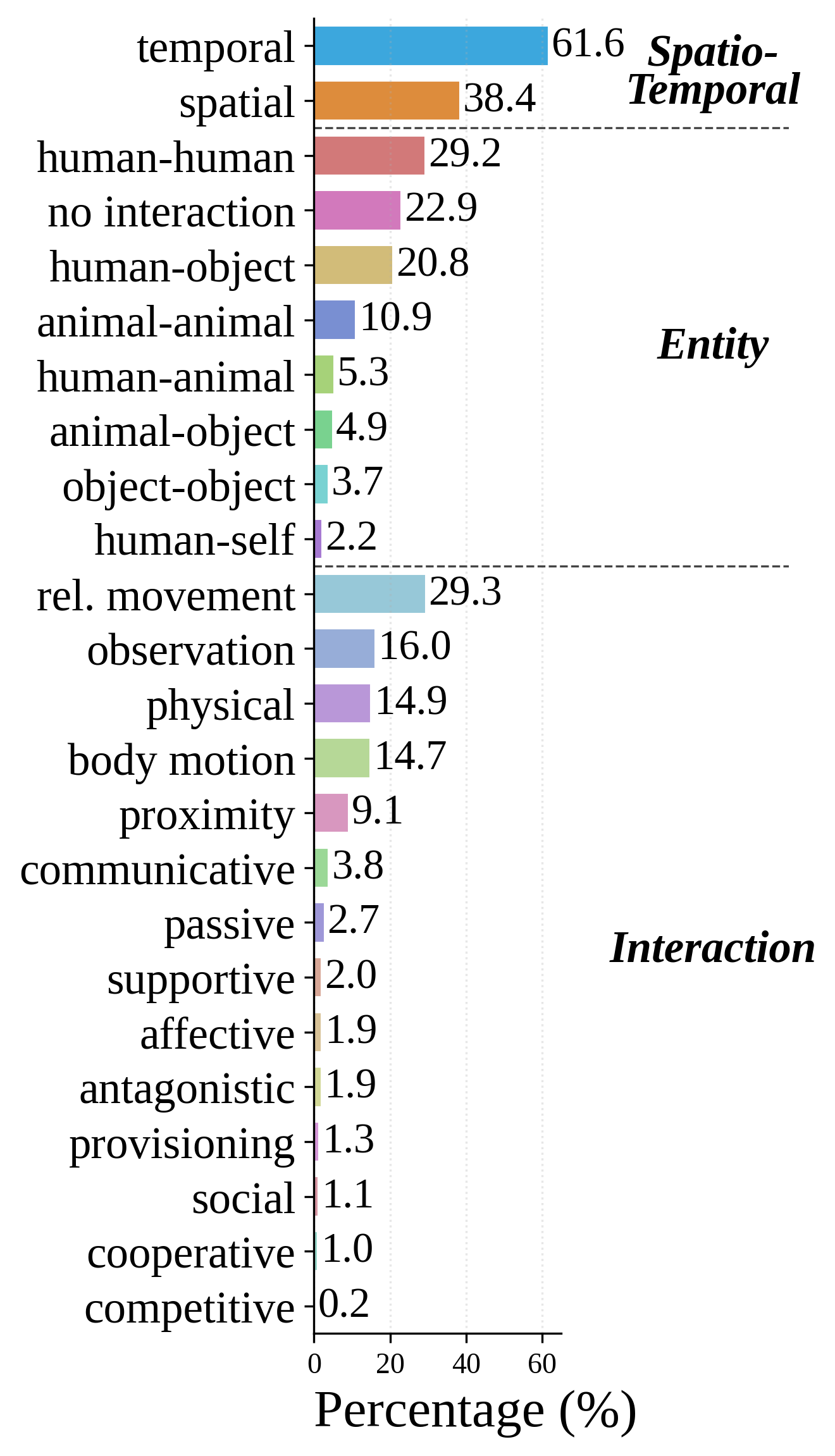}
        \caption{Taxonomy class distribution}
        \label{fig:sub:class_distribution}
    \end{subfigure}%
    \hfill
    \begin{subfigure}[b]{0.48\columnwidth}
        \centering
        \includegraphics[width=\linewidth]
        {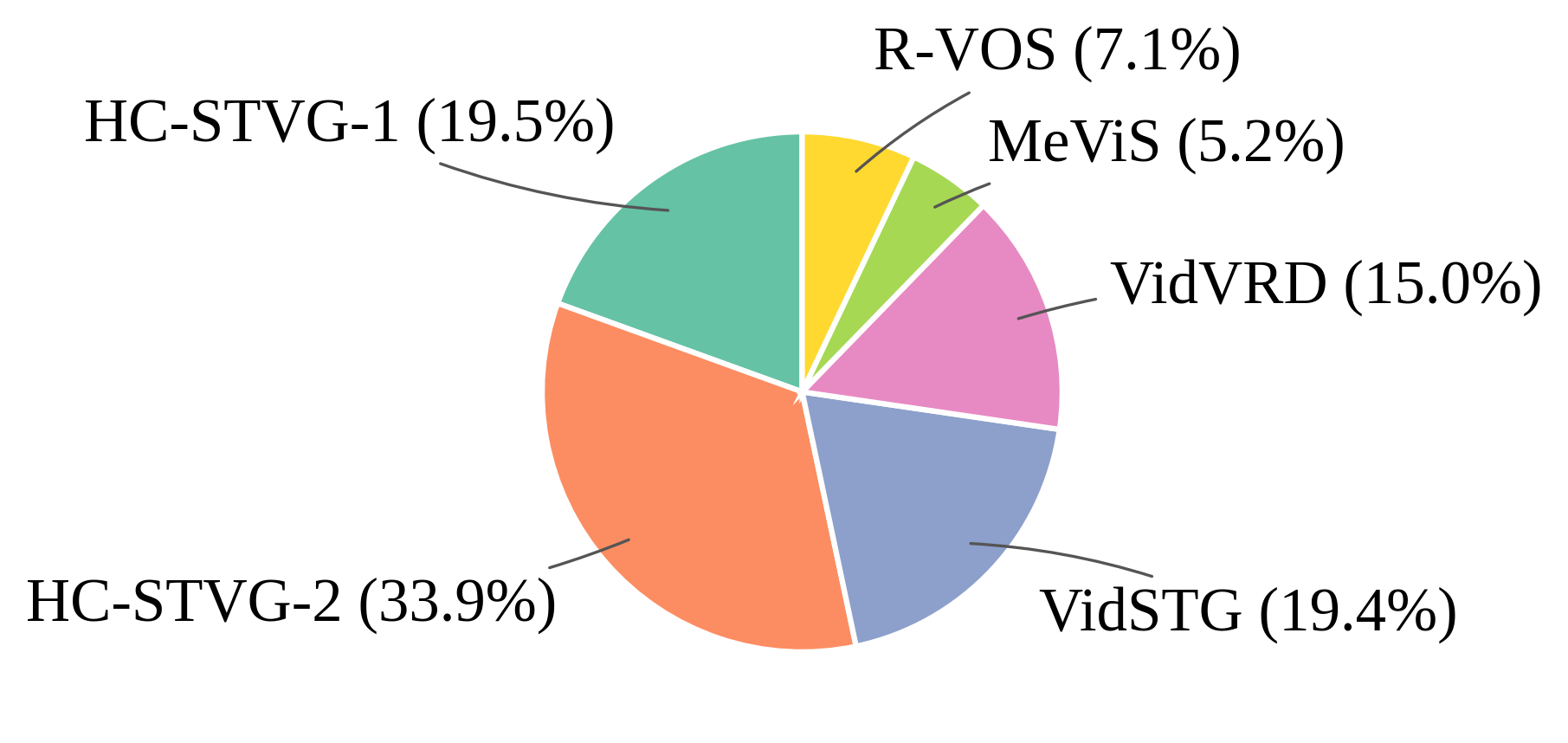}
        \caption{Distribution by dataset}
        \label{fig:sub:datadistribution}
        \vspace{1mm}
        \vspace{1mm}
        \includegraphics[width=\linewidth]
        {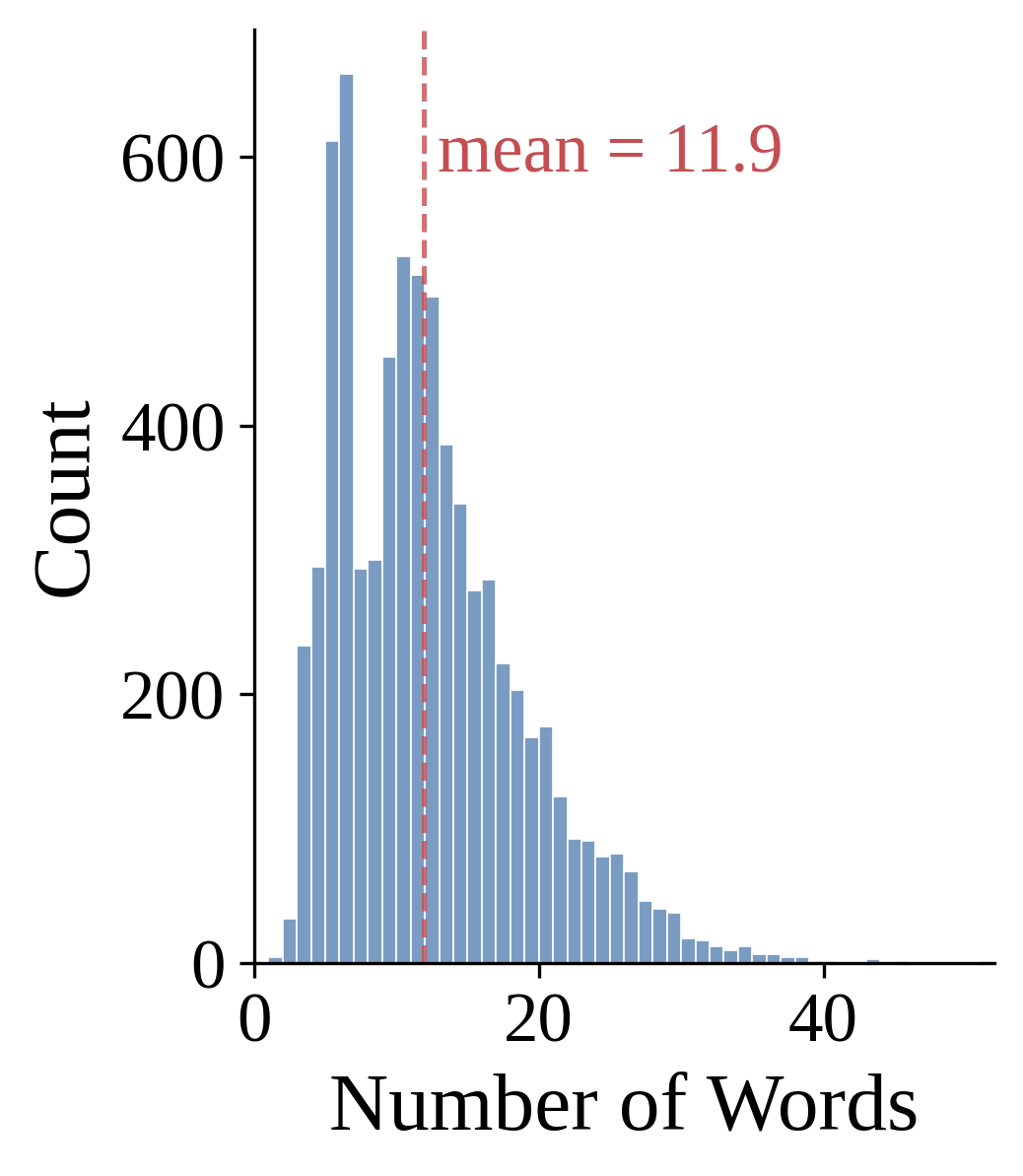}
        \caption{Distribution of caption lengths}
        \label{fig:sub:caption_distribution}
    \end{subfigure}
 
    \caption{Statistical analysis of VISTA Benchmark.}
    \label{fig:VISTA_stats_analysis}
\end{figure}

\noindent \textbf{Motivation:} Despite extensive evaluation of VLMs on spatio-temporal tasks, model failure modes remains largely a black box: aggregate metrics conflate failures across fundamentally different understanding demands, making it impossible to distinguish whether a model struggles with entity identification, spatial grounding, or temporal reasoning. Our taxonomy addresses this by breaking down evaluation into structured, interpretable categories across two complementary levels: \textbf{coarse-grained}, capturing \textit{who} interacts and \textit{where} interactions unfold across space and time, and \textbf{fine-grained}, characterizing the specific relational and behavioral dynamics observed in daily activities. Critically, by stratifying performance across taxonomy categories rather than reporting a single aggregate score, consistent failure patterns and systematic model tendencies become directly visible. \\
\textbf{Coarse-grained} analysis comprises two axes: \textbf{(a) Involved Entities} categorizes interactions based on the involved participants among \textit{humans (H)}, \textit{animals (A)}, and \textit{objects (O)}, capturing all six pairwise configurations: \textit{HH, HA, HO, AA, AO, OO}, augmented with \textit{Human-Self (HS)} for solitary actions and \textit{No Interaction (NO)}. \textbf{(b) Spatio-Temporal Interaction} classifies samples by their primary understanding demand: \textbf{\textit{spatial}} samples focus on positional configurations among entities (e.g., \texttt{"the person beside the car"}), while \textbf{\textit{temporal}} samples capture entity state transitions over time (e.g., \texttt{"the woman sitting down after standing"}). \\
However, coarse categories alone cannot capture the semantic diversity within each bucket - a Human-Human, spatial query may demand relative-position understanding (e.g., \texttt{“the man standing behind the woman"}) or social understanding (e.g., \texttt{"the person comforting the other"}), distinctions a flat taxonomy cannot surface. \textbf{Fine-grained} analysis addresses this across three thematic groups: \textbf{(a) Emotional and Social}: \textit{Affective (AFF)}, \textit{Social (SOC)}, and \textit{Supportive (SUP)} capture emotion, bonding, and assistance; \textbf{(b) Physical and Action-Oriented:} \textit{Physical (PHY)}, \textit{Relational Movement (RM)}, \textit{Cooperative (COP)}, \textit{Competitive (CMP)}, and \textit{Antagonistic (ANT)} describe contact, motion, and joint or opposing effort; \textbf{(c) Observational and Passive:} \textit{Observation (OBS)}, \textit{Communicative (COM)}, \textit{Proximity (PRX)}, \textit{Body Motion (BM)}, \textit{Provisioning (PRV)}, and \textit{Passive (PAS)} reflect non-contact, attention, and static states - together spanning the spectrum of social, physical, and cognitive behavior. The complete taxonomy class distribution can be seen in \autoref{fig:sub:class_distribution}.

\begin{figure}[t!]
    \centering
    \includegraphics[width=\columnwidth]{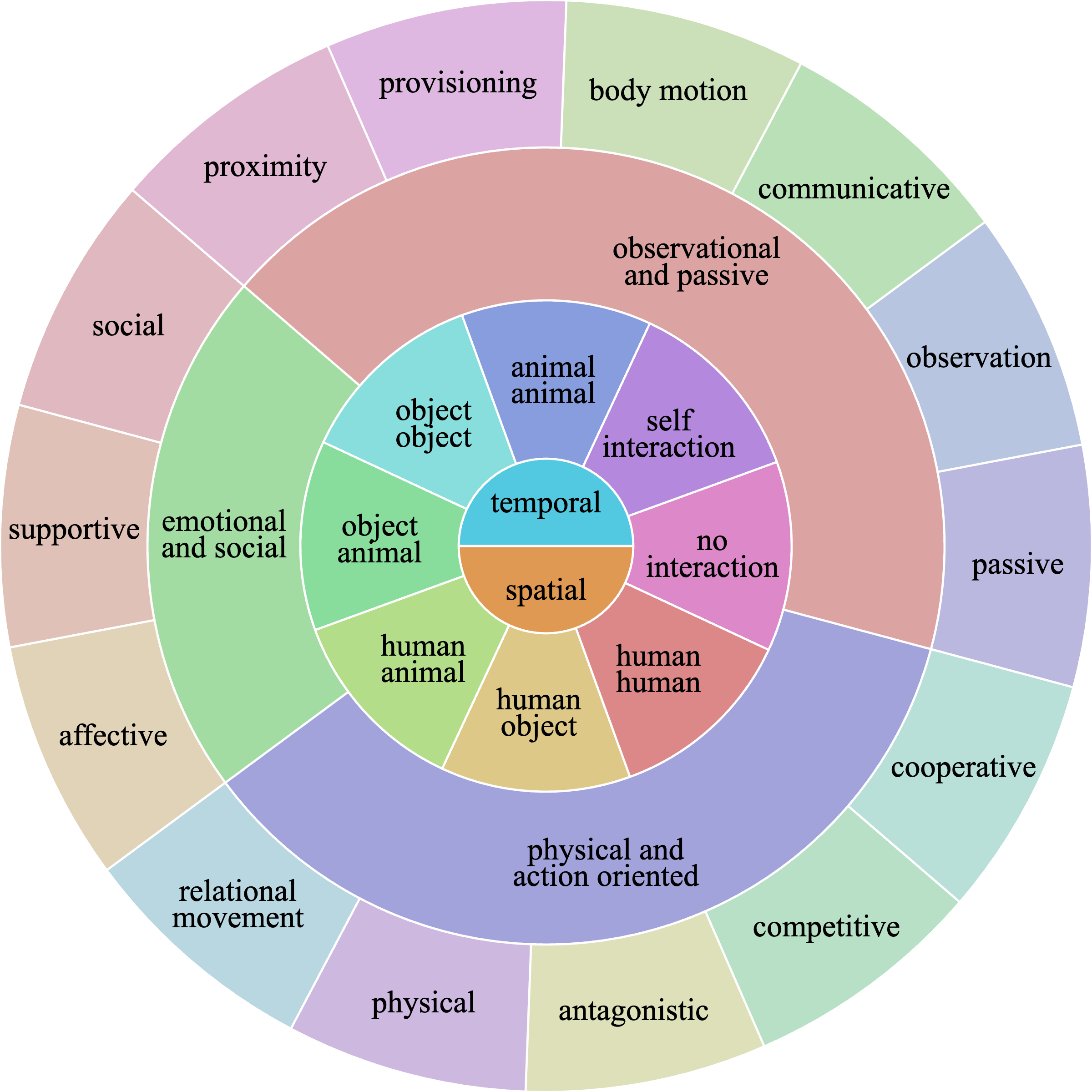}
    \caption{\textbf{Taxonomy} of VISTA benchmark. The two inner circles represent coarse-grained categories, while the outermost circle illustrates the distribution of fine-grained categories.}
    \label{fig:taxonomy}
\end{figure}
\begin{figure*}[t!]
    \centering
    \includegraphics[width=\linewidth]{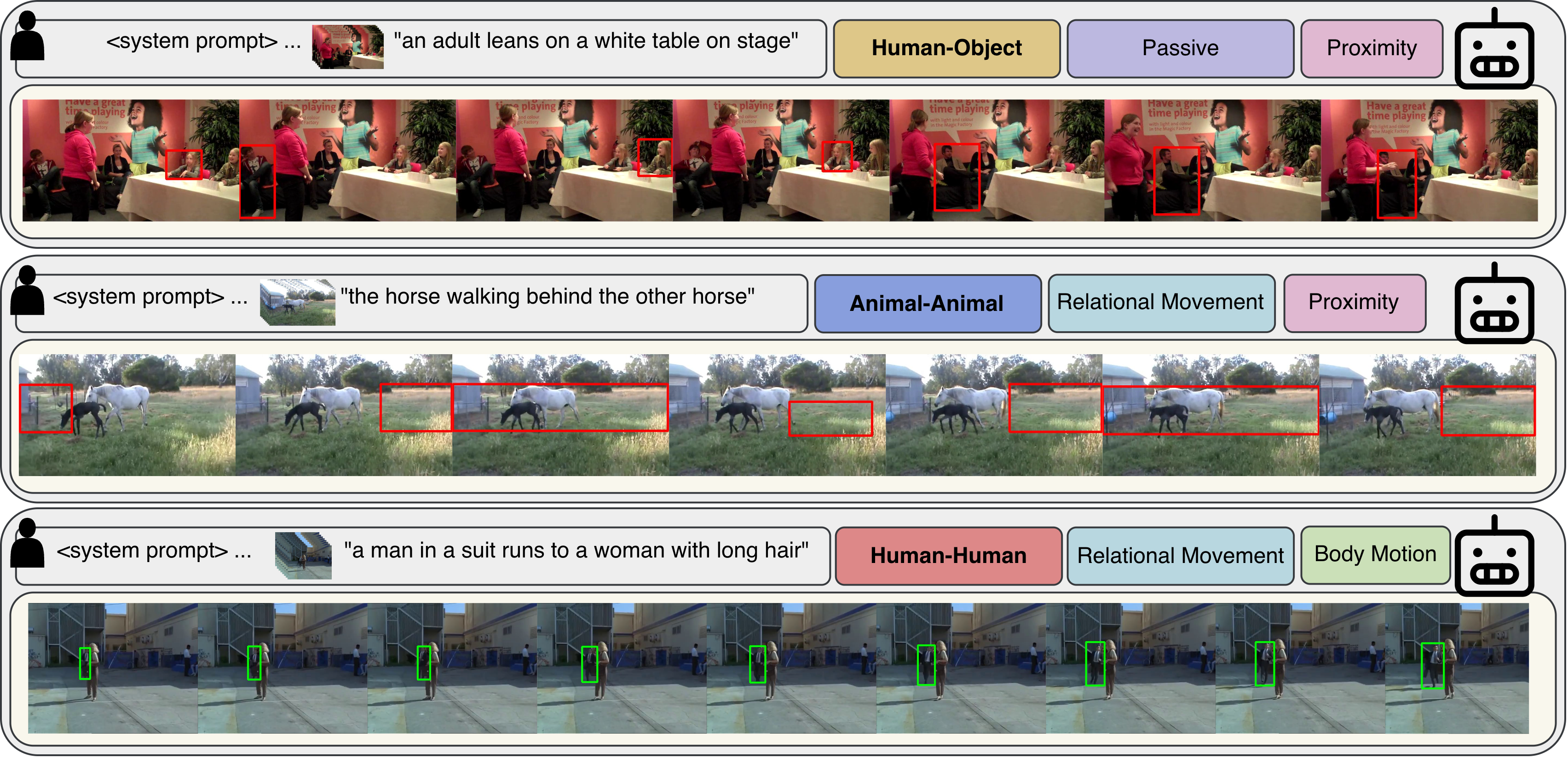}
    \caption{Examples of \textcolor{green}{good} (mvIoU $>0.8$) and \textcolor{red}{bad} (mvIoU $<0.4$) spatio-temporal grounding capabilities across VISTA on the best performing model: CogVLM.}
    \label{fig:qualitative_results}
\end{figure*}
\subsection{Dataset Collection}
\label{sec:dataset}
\noindent \textbf{Motivation:} Prior benchmarks~\cite{zhou2025vlm4d, xu2025mcbench, yuan2025videorefer} have closed object and action vocabularies that restrict evaluation to predefined categories, and templated queries~\cite{zhou2025vlm4d, yuan2025videorefer} capturing only single-step facts that fail to probe the compositional, multi-entity interactions characteristic of real-world video. Our dataset aggregation addresses both - spanning from simple, well-known concepts~\cite{vidvrd, rvos} to fully open-world, complex relational queries~\cite{hcstvg, vidstg, mevis}, with expression styles ranging from template-based to freeform. \\
\noindent \textbf{Dataset Curation:} To build a comprehensive benchmark covering the complexity of spatio-temporal understanding, we aggregate and reformulate six datasets: HCSTVG-v1 and v2 \cite{hcstvg}, VidVRD \cite{vidvrd}, VidSTG \cite{vidstg}, MeViS \cite{mevis}, and RVOS \cite{rvos}. From a language perspective, these datasets span diverse query lengths, reasoning complexity, and expression styles (template-based to freeform). Visually, these datasets span a wide variety of scenes encompassing diverse environments, perspectives and visual challenges such as camera motion, occlusion, and complex object interactions.

\noindent \textbf{Query Formulation:} A core component of our benchmark is the explicit evaluation of human-style narrative queries (freeform) versus template-based queries (referral). Freeform queries capture open-ended, conversational descriptions, while referral queries focus on concise, object-centric expressions.

\begin{itemize}
\item \textbf{Freeform Queries ($Q_F$):} We use freeform captions provided by datasets directly, or reformulate relation triplets (e.g.,  ⟨subject, predicate, object⟩) into freeform natural language sentences through LLMs. Freeform queries capture the full activity and relational context, e.g., \texttt{"A man in a suit walks into the room and sits down"}. 
\item \textbf{Referral Queries ($Q_R$):} Derived by prompting an LLM to extract the primary subject and its attributes from a freeform query. Using the same example, \texttt{"A man in a suit walks into the room and sits down"} reduces to \texttt{"A man in a suit"} - retaining only entity identity and attributes, discarding relational and temporal context entirely.
\end{itemize}

\noindent \textbf{Sample Annotation Pipeline:} For taxonomy classification, we focus exclusively on freeform captions ($Q_F$), which contain the complex relational and spatio-temporal descriptions necessary to assign meaningful interaction categories. We employ a multi-stage pipeline leveraging \texttt{gpt-4o-mini} to classify each caption $qf\in Q_F$ - assigning a single coarse category for involved entities and spatio-temporal interaction, while annotating fine-grained categories exhaustively due to caption complexity. A manual review round was conducted after each classification step to verify and refine labels. Additional implementation details are provided in the supplementary material. 

\noindent \textbf{Annotation Quality:} To validate annotation pipeline reliability, we conducted an inter-annotator agreement study on $n=113$ stratified samples using 2 human annotators and \texttt{gpt-4o-mini}. Cohen's $\kappa$ scores are reported for all three taxonomy levels below.

{\small
\begin{center}
\label{tab:iaa}
\begin{tabular}{@{}lcc@{}}
\toprule
\textbf{Level} & \textbf{H-H $\kappa$} & \textbf{H-GPT $\kappa$} \\ 
\midrule
Entity & 0.98 & 0.76 \\
Spatio-Temporal & 0.77 & 0.69 \\
Fine-grained & 0.83 & 0.67 \\
\bottomrule
\end{tabular}
\end{center}
}

Human-human agreement ($\kappa=0.77-0.98$) indicates substantial to almost perfect agreement~\cite{Landis&Koch}, confirming the taxonomy is well-defined and consistently interpretable across annotators. Human-GPT agreement is moderate ($\kappa=0.67-0.76$), with discrepancies concentrated in visually ambiguous or linguistically underspecified captions - for instance, \texttt{"bear cubs in tow, big bear crossing road"} (GPT: Human-Animal, Corrected: Animal-Animal) and \texttt{"fat man takes out his gun"} (GPT: No Interaction, Corrected: Human-Object). These errors directly motivate the manual verification step in our annotation pipeline. \\
\noindent \textbf{Benchmark Stats:} 
Our benchmark comprises 11,814 unique video–caption pairs $(V, Q)$, offering a rich set of fine-grained annotations. Textual descriptions range between 40-60 words on average, reflecting the complexity of the freeform language used. Video resolution and number of frames are approximately $866\times544$ pixels and $174$ frames, respectively. This combination of detailed spatio-temporal annotations, realistic video lengths, and varied scene content distinguishes our benchmark from existing datasets. The distributions by datasets and description length are illustrated in \autoref{fig:sub:datadistribution} and \autoref{fig:sub:caption_distribution}, respectively. The fine-grained distribution reflects the organic frequency of interaction types in natural video: Competitive (0.2\%) and Cooperative (1.0\%) are genuinely rare while Relational Movement (29.3\%) and Observation (16.0\%) are not. Fine-grained analysis spans all categories for diagnostic breadth; quantitative conclusions are restricted to categories with sufficient sample support.

\subsection{Evaluation Setup} 

\noindent \textbf{Benchmark Models:} 
Building on prior work~\cite{xie2023DOD}, we select a representative set of models capturing diversity in architecture (LLM-based vs. non-LLM-based), training paradigm, and task specialization, as these factors naturally influence grounding capabilities. A fundamental requirement for inclusion is the ability to produce structured bounding-box predictions necessary for IoU-based evaluation - while powerful model families like GPT, Gemini, and VideoLLaMA demonstrate spatio-temporal understanding, they are not explicitly trained for fine-grained localization, making reliable IoU-based assessment infeasible. We organize selected models into three categories: \textbf{(1) Foundation Models without LLMs}, \textbf{(2) Generalist MLLMs}, and \textbf{(3) Specialist MLLMs}.
Category (1) includes Grounding-DINO~\cite{Liu2023GroundingDM} for its strong zero-shot detection generalizability. Category (2) comprises Intern-VL 2.5~\cite{internvl}, Mini-GPT-v2~\cite{chen2023minigptv2}, Sphinx-v2~\cite{sphinx-mllm}, Qwen-VL-Chat~\cite{qwenvl-mllm}, and Qwen3-VL~\cite{qwen3vl} - LLM-backed models trained on diverse tasks including localization. Category (3) represents task-specific models trained exclusively for detection and related localization tasks and includes Shikra~\cite{shikra}, Ferret~\cite{ferret}, and CogVLM~\cite{cogvlm} which generate bounding boxes in plain text, and LLaVA-Grounding~\cite{zhang2024llavag} which combines a LLM with a dedicated detection head. All models are evaluated zero-shot on sub-sampled video frames. Further details are in the supplementary.

\noindent\textbf{Model Selection Rationale.}
We prioritize open-weight models for two reasons: (1)~\emph{Reproducibility} - proprietary models such as GPT-4o undergo silent updates that can substantially alter behavior between evaluation runs~\cite{chen2023chatgpt}, undermining the diagnostic consistency central to VISTA's contribution; and (2)~\emph{Cost} - systematic evaluation across {$\sim$}12K video--query pairs with multi-frame sampling is prohibitively expensive through commercial APIs. We note that VISTA's evaluation framework and taxonomy are model-agnostic and directly applicable to proprietary or future models as access constraints evolve. 

\noindent\textbf{Data Contamination:} We cross-referenced all VISTA video identifiers against the disclosed training splits of all evaluated models, finding no overlaps. Full decontamination remains infeasible given incomplete disclosure of web-scale pretraining corpora; however, our analyses focus on intra-model performance stratification rather than absolute scores. Relative patterns such as the cross-entity vs. same-entity gap are robust to incidental exposure, as contamination inflates scores uniformly across categories.

\noindent \textbf{Evaluation Metrics:} We report performance using metrics established in previous studies \citep{Yang2022TubeDETRSV, Jin2022EmbracingCA, kumar2025contextual, garg2025stpro}: mean spatio-temporal IoU ($m\_vIoU$) which is computed as $\frac{1}{|S_{u}|} \sum_{t\in S_{i}}$IoU$(\hat{b_{t}}, b_{t})$, where $S_{i}$ and $S_{u}$ denote the intersection and union, respectively, between the predicted and ground truth timestamps. IoU$(\hat{b}_t, b_t)$ represents the spatial overlap between the predicted bounding box $\hat{b}_t$ and the ground truth box $b_t$ at frame $t$. 

\section{Directional Biases in Interactions}
We evaluate and analyze the relative performance of models across \textbf{VISTA}'s 
hierarchical taxonomy, examining differences both intra and inter model families. 
Our analysis proceeds along three axes: query structure (referral vs. freeform), 
coarse-grained analysis, fine-grained analysis. Across this taxonomy, several trends emerge. Model performance follows a clear family-level hierarchy, with Generalist MLLMs outperforming both Specialist MLLMs 
and Foundation Models. More notably, models exhibit a consistent sensitivity to 
query structure, performing substantially better on referral than freeform queries 
across all families - indicating continued reliance on syntactic scaffolding over 
genuine multimodal reasoning. At the interaction level, same-entity interactions 
reveal systematic symmetry failures, while the relatively balanced performance 
across spatial and temporal samples stands in contrast to the static, image-based 
nature of most model training. Beyond these trends, we examine the directional 
tendencies underlying these failures - reasoning about how pretraining distributions, 
cross-modal alignment, and architectural choices systematically shape model 
interpretation. Nearly all reported performance differences are statistically significant 
($p<0.05$); bootstrap confidence intervals and full hypothesis test details are provided in the supplementary.

\subsection{Impact of Query Structure}
\label{sec:query-structure}
\autoref{tab:zs_ref_vs_free_det}, reveals a robust and repeatable pattern across all model families: \hl{\textit{models perform better on referral (template-like) queries than free-form (natural language) queries}}. This gap indicates that models are sensitive to prompt structure - leveraging syntactic cues such as \texttt{<subject, verb, object>} ordering when present, but failing to compensate through multimodal reasoning when they are not. This failure mode is exemplified in \autoref{fig:qualitative_results} (top), where, given the query \texttt{``an adult leans on a white table on stage''} the model successfully grounds \texttt{``an adult''} but fails to leverage the spatial cue \texttt{``on the table''} to recognize that the subject is actually a child. This highlights how models prioritize syntactic patterns over spatial reasoning cues embedded in natural language. More broadly, pre-training breadth shapes this gap directly: models trained on heterogeneous, interaction-rich corpora maintain more stable R-F balance, while those fine-tuned on narrow domains or static captions degrade under freeform settings, overfitting to surface co-occurrence statistics rather than learning compositional reasoning. A notable exception is Qwen3-VL, the strongest model overall, which reverses this trend with freeform queries (64.41) outperforming referral queries (62.85), suggesting that sufficient pretraining breadth and instruction diversity can enable models to exploit richer context in freeform descriptions rather than relying on syntactic scaffolding. MimoVL achieves competitive generalist performance but exhibits a pronounced same-entity deficit (OO: 27.0 vs. HO: 46.0), consistent with the broader disambiguation failures identified across models. A subset of Specialist MLLMs exhibit marginal gains on freeform inputs, indicating that \hl{\textit{additional linguistic context can be beneficial when it aligns with a model's training distribution}}.

\subsection{Coarse Grained Category Analysis}
\label{sec:coarse-grained-analysis}
Coarse-grained interactions are analyzed along two axes: involved entities and 
spatio-temporal interaction type as per taxonomy sub-divisions.

\begin{figure*}[t!]
    \centering
    \includegraphics[width=\textwidth]{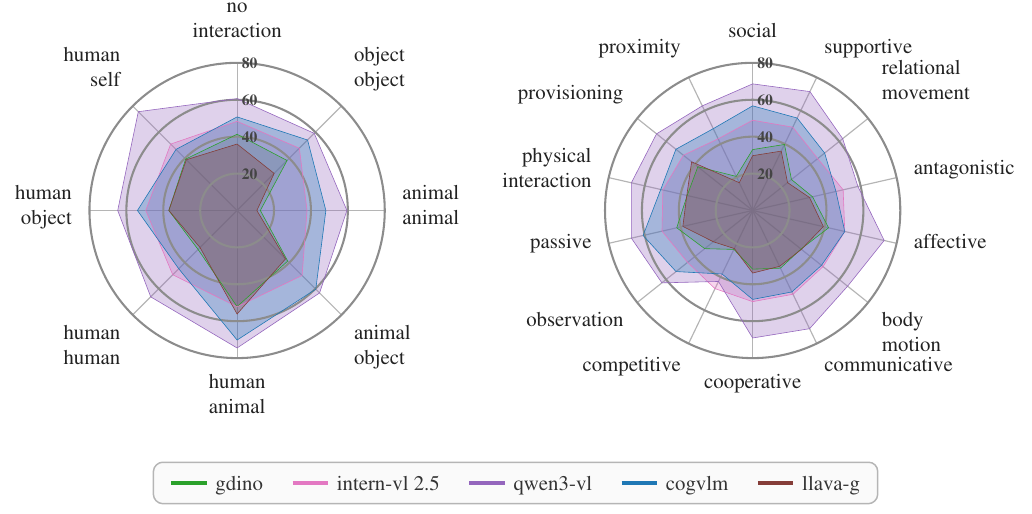}
    \caption{Per-model mvIoU across (left) coarse-grained entity-pair categories and (right) fine-grained interaction types. Cross-entity pairs (e.g., Human-Animal) consistently outperform same-entity pairs (e.g., Animal-Animal), while interactions with salient visual cues (e.g., relational movement) yield stronger performance than those requiring implicit reasoning (e.g., passive, affective).}
    \label{fig:fine_grained_radar}
\end{figure*}

\noindent \textbf{Involved Entities:} \autoref{fig:fine_grained_radar}(a) reveals a clear 
pattern across models: \hl{\textit{interactions that cross entity categories score 
substantially higher than same-entity interactions}}. Averaging across models, 
\textit{Human-Animal (HA) interactions are the strongest (51.6\% avg.)}, while 
\textit{Animal-Animal (AA) interactions are the weakest (31.1\% avg.)}, and 
Object-Object (OO) interactions are also relatively low (37.3\% avg.). This 
reflects a prevalent failure mode rooted in \emph{category-level priors}: models 
can more effectively ground entities when they belong to different semantic classes, 
but struggle to disambiguate visually similar instances of the same class, instead 
defaulting to general entity recognition rather than leveraging specific referential 
cues. This pattern persists even in high-performing Generalist MLLMs, indicating 
that representational homogeneity, rather than limited capacity, drives these errors.
\autoref{fig:qualitative_results} (middle) illustrates this symmetry failure 
explicitly: given the query \texttt{``the horse walking behind the other horse''}, 
CogVLM fails to resolve the spatial relationship \texttt{``behind''}, defaulting 
to grounding an entity of the correct class rather than the specific one requested. 
This contrasts with \autoref{fig:qualitative_results} (bottom), where given 
\texttt{``a man in a suit runs to a woman with long hair''}, the model successfully 
grounds \texttt{``the man in a suit''}. Although this is also a same-entity 
(Human-Human) interaction, the entities are visually distinct and described by 
their attributes (\texttt{``in a suit,''} \texttt{``with long hair''}) rather than 
a complex spatial relation - confirming that the core failure lies in lack of reasoning 
about relational and spatial cues when visual distinctiveness between entities is low.

\noindent \textbf{Spatio-Temporal Interaction:} \hl{Performance across spatial (S) and 
temporal (T) samples is roughly comparable across all model families, 
\textit{suggesting no strong global bias for one axis}} -- a finding that runs 
counter to expectations, given the predominantly static, image-based training of 
most architectures. Examining this by model family reveals an interesting split: 
foundation and specialist models conform to the expected spatial bias, remaining 
anchored to static appearances consistent with their training. LLM-based generalist 
models exhibit near-parity between spatial and temporal performance, suggesting that 
jointly decoding over language and vision features helps compensate for challenging 
temporal visual conditions such as motion blur or occlusion. 

These coarse-level trends highlight two 
complementary limitations in current VLMs. First, grounding performance is strongly 
conditioned on visual and semantic distinctiveness between entities - models succeed 
when category or appearance differentiates the referent, but fail when disambiguation 
requires relational or spatial reasoning. Second, while surface-level spatial and 
temporal performance appears balanced, this masks an underlying preference for static 
configurations: models struggle with causality, motion sequences, and transitions 
in visually ambiguous scenes.

\subsection{Fine Grained Category Analysis}
\label{sec:fine-grained-analysis}
Fine-grained interactions reveal deeper insights into how models handle interpersonal, physical, and non-contact nuances beyond coarse entity and space–time reasoning. \autoref{fig:fine_grained_radar}(b) shows that Generalist MLLMs consistently achieve the highest scores, while Specialist MLLMs and Foundation Models exhibit sharp variability depending on the type of interaction. A key trend is that \hl{\textit{models perform substantially better on interactions with clear visual anchors}} (e.g., physical interaction, supportive, social) and struggle when the interaction requires \hl{\textit{implicit cognition, emotional inference, passive states}}.

\textit{(a) Emotional and Social:} Models show moderate performance on affective, social, and supportive interactions, yet these categories remain consistently among the weakest across all model families, indicating that \hl{\textit{MLLMs lack robust grounding for subtle emotional or interpersonal behaviors}}, particularly when cues are indirect or language-driven. This weakness is compounded by a broader tendency we term \emph{semantic-intent inflation}: instruction-tuned and generalist MLLMs systematically over-interpret scenes through high-intent or affective frames, projecting social and emotional significance onto interactions even when the visual evidence supports simpler physical or positional readings. \autoref{fig:qualitative_results} (top) illustrates this directly: given \texttt{``an adult leans on a white table on stage,''} the model grounds \texttt{``an adult''} but fails to leverage the spatial cue \texttt{``on the table''} to recognize that the subject is a child. Rather than parsing the \textbf{Proximity (PRX)} and \textbf{Passive (PAS)} nature of the interaction, the model defaults to a socially inflated interpretation anchored in the referral term. This pattern reflects pretraining distributions and instruction tuning that emphasize conversational and affective content, systematically biasing models toward semantic over-attribution even at the cost of spatial and relational accuracy.

\textit{(b) Physical and Action-Oriented:} These interactions yield the strongest performance overall, particularly for generalist models, as they involve motion, contact, or clear physical consequences that provide salient visual anchors. Yet even within this group, important distinctions emerge: cooperative and physical interactions benefit from visually structured cues, while \hl{\textit{competitive and antagonistic actions remain harder to disambiguate}}, requiring models to distinguish between semantically similar but directionally opposed dynamics. Moreover, kinematic and relational reasoning breaks down even when the broader category is favorable. \autoref{fig:qualitative_results} (middle) illustrates this: given \texttt{``the horse walking behind the other horse,''} CogVLM fails to parse \texttt{``behind''} as a \textbf{Relational Movement (RM)} relationship, defaulting to entity recognition rather than modeling the directional dynamic between two visually similar entities. This failure reveals that strong aggregate performance on physical interactions masks a specific deficit in directional and motion-based reasoning. \hl{\textit{Models can leverage visual salience when interactions produce observable consequences}}, but struggle when grounding depends on parsing the spatial trajectory or relative motion between entities rather than identifying the entities themselves.
 
\textit{(c) Observational and Passive:} Performance splits sharply within this group. Interactions with explicit visual cues, such as proximity and body motion, are handled reasonably well, whereas passive or cognitive categories such as observation and provisioning remain challenging, as they require inferring intent, attention, or perspective from subtle or absent visual signals. This difficulty exposes a systematic tendency we term \emph{social-first bias}: when an interaction contains both social and physical signals, models interpret it primarily through the lens of identity and affect, often at the expense of physical dynamics. \autoref{fig:qualitative_results} (bottom) exemplifies this: given \texttt{``a man in a suit runs to a woman with long hair,''} the model successfully grounds the correct entity, but its success stems from over-reliance on distinctive textual attributes (\texttt{``in a suit,''} \texttt{``with long hair''}) rather than genuine understanding of the underlying \textbf{Body Motion (BM)} and \textbf{Relational Movement (RM)}. The model treats the interaction as an identity-matching problem rather than a kinematic one. This shallow grounding strategy generalizes across architectures: \hl{\textit{even when models produce correct predictions on observational or passive interactions, the reasoning pathway frequently bypasses the physical and attentional cues}} that define the category. \\
Across all three fine-grained groups, a consistent pattern emerges: current VLMs are more adept at reasoning about \emph{why} interactions occur than \emph{how} they unfold or \emph{where} they are situated. Semantic-intent inflation and social-first bias are complementary manifestations of the same underlying gap. Alignment strategies, instruction tuning, and pretraining distributions have successfully taught models to emphasize semantic content, particularly social and affective aspects, but have not sufficiently reinforced the modeling of physical motion, relational dynamics, or subtle spatial states. Correcting this imbalance will require integrating datasets with complex kinematic cues and multi-agent dynamics, alongside explicit grounding tasks that force models to jointly reason about social intent, temporal evolution, and spatial configuration.

\section{Conclusion}

In this work, we introduce \textbf{VISTA}, a benchmark for evaluating fine-grained, interaction-centric spatio-temporal reasoning in Vision-Language Models (VLMs). By unifying multiple datasets into a single interaction-aware taxonomy and decomposing videos into coarse- and fine-grained interaction types, VISTA enables a critically nuanced evaluation of spatial, temporal, and relational understanding. Our framework not only exposes hidden model weaknesses masked by aggregate metrics but also characterizes generalization patterns and uncovers directional, spatial, and temporal biases across a broad range of state-of-the-art models. Through extensive experiments over several modern MLLMs, we demonstrate that even high-performing models exhibit limitations in multi-entity, multi-action, and temporally compositional reasoning, with same-entity disambiguation and semantic-intent inflation emerging as the two most critical bottlenecks. These findings suggest that targeted training on visually similar multi-instance scenes and kinematic reasoning tasks may yield the largest gains. Overall, VISTA provides a first systematic lens for diagnosing these limitations, bridging the gap between abstract-level assessment and robust, real-world video understanding.

{
    \small
    \bibliographystyle{ieeenat_fullname}
    \bibliography{main}
}

\end{document}